\documentclass[11pt]{article}

\usepackage[margin=1in]{geometry}
\usepackage{graphicx}
\usepackage{float}
\usepackage{caption}
\usepackage{subcaption}
\usepackage{amsmath,amssymb,amsfonts}
\usepackage{booktabs}
\usepackage[hidelinks]{hyperref}
\usepackage{placeins}
\usepackage{authblk}
\usepackage{microtype}
\usepackage{xcolor}
\usepackage{tikz}
\usetikzlibrary{arrows.meta,positioning,calc,decorations.pathreplacing}

\definecolor{draftred}{RGB}{135,25,25}
\definecolor{rfblue}{RGB}{45,115,175}
\definecolor{dforange}{RGB}{220,115,35}
\definecolor{softgray}{RGB}{244,244,244}

\newif\ifdraftnotes
\draftnotesfalse
\newcommand{\currentRuns}{1000}
\newcommand{\currentSteps}{200{,}000}
\newcommand{\currentBurnIn}{40{,}000}

\title{\texorpdfstring{%
\textbf{Intermittent Control Is Not Diluted Control:}\\[0.4em]
A Switching Effect in Artificial Agency%
}{Intermittent Control Is Not Diluted Control: A Switching Effect in Artificial Agency}}
\author[1]{Veronique Ziegler}
\affil[1]{Independent Researcher}
\date{}

\begin{document}
\maketitle

\begin{abstract}
Adaptive agents do not always regulate under the same timing conditions. Sometimes stabilization can begin before a disturbance has fully entered the internal state; at other times, the agent can only recover after disruption has already taken hold. A simple expectation is that an agent moving between these two conditions should behave like a weighted average of the two fixed cases: the more time spent in reactive recovery, the greater the regulatory burden.

This paper shows that this expectation can fail. In a simulated adaptive agent with retained state history, and at an operating point where sustained reactive control is more costly than sustained anticipatory control, intermittent access to anticipatory control reduces the mean regulatory burden below the value predicted by a fixed-mode mixture of the two. The effect appears under both periodic and stochastic switching schedules: losing anticipatory access does not simply dilute its benefit, and restoring it intermittently can reorganize the later regulatory burden. High-statistics runs ($N=1000$ matched replicates per schedule) resolve a negative nonlinear switching penalty across every tested schedule. The effect is small but consistent: about half a percent of the mean gain, with 63--68\% of individual replicates falling below zero. Late-window diagnostics reveal no unresolved upward accumulation of regulatory burden.

The result identifies a design-relevant timing principle for artificial agents. In history-dependent adaptive systems, the burden of remaining organized is not determined only by how much time an agent spends in anticipatory or reactive mode. The order in which disturbance and recovery enter the state can change the subsequent regulatory burden. Intermittent anticipatory control may therefore act less like a partial failure of regulation and more like a mechanism for reducing the long-term burden of recovery.
\end{abstract}

\section{Introduction}

Artificial agents are usually evaluated under simplified assumptions about operating mode~\cite{russell2020aima,sutton2018rl}. A controller is tested with a given disturbance level, a policy is compared with another policy, or a recovery mechanism is evaluated after a fixed failure. These comparisons are useful, but they can miss a basic feature of realistic adaptive operation: access to anticipatory regulation may itself be intermittent. A robot may briefly lose reliable sensing, a monitoring agent may experience delayed observations, an autonomous planner may operate under fluctuating compute resources, and a simulated adaptive agent may alternate between predictable and disrupted environments. In such cases, the agent does not live permanently in one causal regime. It repeatedly loses and regains the ability to regulate before disturbance has strongly altered its internal state.

The simulation presented here tests whether that alternation changes the regulatory burden. If an agent spends half of its analyzed time in an anticipatory regime and half in a reactive regime, should its long-run regulatory burden equal the average of the two fixed regimes? That expectation treats switching as a bookkeeping problem: count the RF time, count the DF time, and average the fixed baselines. But adaptive agents retain state history. Their internal uncertainty, controller gain, and subsequent susceptibility can be changed by previous exposure. For that reason, the same amount of reactive operation may have a different cost depending on how it is embedded in time.

The present study examines this question in IRAM-$\Omega$-Q, a compact computational model of uncertainty regulation in artificial agents~\cite{ziegler2026iramomegaq}. It uses a quantum-like state representation, in the spirit of quantum-like models of cognition and decision~\cite{busemeyer2012quantum}, because density matrices provide direct observables for entropy, coherence, and internal organization. Within that model, an adaptive gain $\mu(t)$ regulates uncertainty relative to a target $S^*$. The gain is interpreted as regulatory burden: higher values mean that the trajectory requires stronger modeled stabilizing intervention to remain organized.

Two causal orderings are compared. In regulation-first (RF) ordering, the controller updates before current disturbance exposure and can attenuate the incoming disturbance. In disturbance-first (DF) ordering, the disturbance enters before a newly computed regulatory response is available. Earlier IRAM-$\Omega$-Q studies held this ordering fixed over a trajectory and found that persistent DF operation requires greater regulation than persistent RF operation under comparable conditions \cite{ziegler2026iramomegaq,ziegler2026carryover}. Those fixed-ordering results motivate the present question: what happens when one agent alternates between the two orderings?

\begin{figure}[t]
\centering
\includegraphics[width=0.93\linewidth]{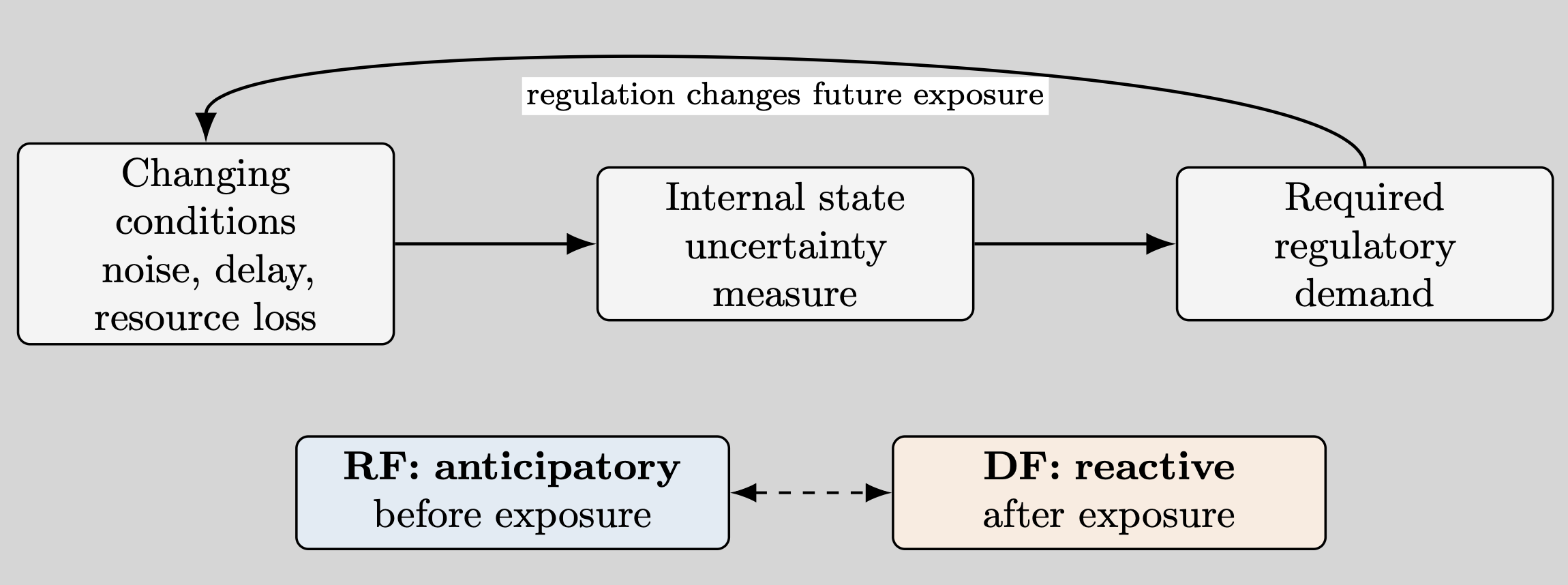}
\caption{\textbf{Broader artificial-agent question.} An adaptive agent may alternate between anticipatory regulation (RF) and reactive regulation (DF) as operating conditions change. The study tests whether switching burden is captured by averaging fixed RF and fixed DF operation, or whether temporal alternation itself produces a nonlinear effect.}
\label{fig:agent_question}
\end{figure}

The original expectation was that intermittent loss of RF would add burden. Reactive intervals could perturb the internal state, and the resulting correction might persist even after RF access returned. This would produce a positive nonlinear switching penalty. Instead, the simulations show a robust negative penalty. Across periodic and stochastic switching schedules, the switching agent requires less mean regulatory gain than predicted by an occupancy-weighted mixture of matched fixed RF and fixed DF reference trajectories. The effect is therefore not merely that RF is better than DF. The stronger result is that intermittent restoration of RF changes the burden of DF exposure relative to fixed-mode averaging (Figure~\ref{fig:agent_question}).

The contribution of this paper is to formulate, measure, and bound this switching effect. Specifically, the study:
\begin{enumerate}
    \item defines intermittent anticipatory access as switching between RF and DF causal orderings in a single adaptive agent;
    \item constructs an occupancy-adjusted fixed-mode reference for testing whether switching is equivalent to averaging fixed RF and DF operation;
    \item demonstrates a robust negative switching penalty in regulatory gain at a high-statistics reference condition;
    \item checks late-time stationarity to reduce the possibility that the effect is only an unresolved finite-time drift;
    \item maps persistence and attenuation over a bounded neighborhood in $(\mu_0,\eta,S^*)$; and
    \item identifies the boundary of the observed effect by comparing switching relief with the local fixed RF-versus-DF regulatory-burden separation.
\end{enumerate}

The result has a deliberately narrow scope. IRAM-$\Omega$-Q is used here as a transparent mathematical testbed for time-varying access to anticipatory regulation. Within this testbed, switching between causal orderings is not reducible to fixed-mode occupancy alone. Because the agent retains internal history, the sequence of disruption and recovery changes the subsequent regulatory burden. This identifies a concrete mechanism for comparison with other adaptive-agent models, rather than a claim of universality.

\section{Model: An Adaptive Agent with Time-Ordered Regulation}
\label{sec:model}

The purpose of the model is to isolate a timing question that is difficult to interpret in a large AI system: how does regulatory burden change when the agent alternates between anticipatory and reactive access to control? IRAM-$\Omega$-Q provides a controlled setting in which internal uncertainty, regulatory gain, causal ordering, and state history are all directly measurable.

\subsection{Operational interpretation}

The agent is represented by an evolving internal state, a target uncertainty, and a feedback controller~\cite{astrom2008feedback}. The model quantities are interpreted operationally rather than anthropomorphically. They describe the behavior of a simulated adaptive system, not subjective experience.

\begin{table}[t]
\centering
\caption{Operational interpretation of the main model quantities used in this paper.}
\label{tab:model_quantities}
\begin{tabular}{p{0.25\linewidth} p{0.65\linewidth}}
\toprule
Model quantity & Operational interpretation in the artificial agent \\
\midrule
Internal state $\rho(t)$ &
Structured internal representation whose organization can be disrupted, stabilized, and carried forward through time. \\

Entropy $S_{\mathrm{vN}}$ &
Modeled internal uncertainty relative to a target operating range. \\

Coherence gap $\Delta C$ &
Diagnostic of organized structure not captured by diagonal uncertainty alone. \\

Gain $\mu(t)$ &
Intensity of adaptive stabilizing intervention required along the trajectory. \\

RF / DF ordering &
Whether updated regulation acts before or after the current disturbance exposure. \\
\bottomrule
\end{tabular}
\end{table}

\subsection{State representation and observables}

The present paper retains the state representation and adaptive-gain architecture introduced in the framework paper \cite{ziegler2026iramomegaq}, together with the regulation-first (RF) and disturbance-first (DF) causal orderings and operational diagnostics developed in the subsequent hysteresis study \cite{ziegler2026carryover}. The definitions below are restated only to fix the quantities used in the switching analysis.

The primary dynamical state is a normalized complex amplitude vector,
\begin{equation}
\psi(t)=\bigl(\psi_1(t),\ldots,\psi_d(t)\bigr)\in\mathbb{C}^{d},
\qquad
\sum_{i=1}^{d}|\psi_i(t)|^2=1.
\end{equation}
At each measurement step, the implementation constructs a density representation $\rho(t)$ from the current state and uses it to compute the controller uncertainty signal and coherence-gap diagnostic.

Consistent with that earlier work \cite{ziegler2026carryover}, the controller's uncertainty signal is computed from this density representation using the von Neumann entropy form,
\begin{equation}
S_{\mathrm{vN}}(\rho)
=
-\mathrm{Tr}(\rho\log\rho)
=
-\sum_k \lambda_k \log \lambda_k ,
\end{equation}
where $\{\lambda_k\}$ are the eigenvalues of $\rho(t)$. Throughout, $S_{\mathrm{vN}}$ is used operationally as the model's feedback signal for internal uncertainty and carries no stronger physical interpretation. The diagonal entropy is
\begin{equation}
S_{\mathrm{diag}}(\rho)
=
-\sum_i \rho_{ii}\log\rho_{ii},
\end{equation}
and the coherence gap is
\begin{equation}
\Delta C = S_{\mathrm{diag}} - S_{\mathrm{vN}} .
\end{equation}

The adaptive gain $\mu(t)$ controls the strength of stabilization and is interpreted here as a model-level measure of regulatory burden. In the present paper, $\mu(t)$ is the primary burden measure, while $\Delta C$ is used as a state-level diagnostic of retained internal organization.

\subsection{Adaptive regulation}

The controller regulates entropy relative to a target $S^*$. With entropy error
\begin{equation}
e(t)=S_{\mathrm{vN}}(t)-S^*,
\end{equation}
the implementation adapts the gain as
\begin{equation}
\mu(t+\Delta t)
=
\Pi_{[\mu_{\min},\mu_{\max}]}
\left[
\mu(t)
+
\alpha\,\dot S_{\mathrm{vN}}(t)
+
\beta\,e(t)
\right],
\label{eq:mu-update}
\end{equation}
where $\dot S_{\mathrm{vN}}$ is a finite-difference entropy-change signal, $\Pi_{[\mu_{\min},\mu_{\max}]}$ denotes clipping to the admissible gain interval, and $\alpha$ and $\beta$ are the derivative and target-error feedback gains listed in Table~\ref{tab:protocol}. With the error convention $e(t)=S_{\mathrm{vN}}(t)-S^*$, entropy above target contributes positively through the $\beta$ term and therefore increases regulatory gain, as the feedback interpretation requires. Equation~\eqref{eq:mu-update} is the update rule of the framework and hysteresis papers \cite{ziegler2026iramomegaq,ziegler2026carryover}, carried over unchanged; higher $\mu(t)$ represents stronger stabilizing intervention demanded by the simulated trajectory.

After ordering-specific disturbance and stabilization operations, coherent internal evolution is applied through the same propagator in both orderings,
\begin{equation}
\psi(t+\Delta t)=U_{H,\Delta t}\psi_{\mathrm{nc}}(t),
\qquad
U_{H,\Delta t}=e^{-iH\Delta t}.
\end{equation}
Thus RF and DF differ in the causal location of regulation relative to disturbance exposure, not in the coherent evolution rule.

\subsection{Fixed RF and DF orderings}

Let $\mathcal{C}$ denote the controller update, $\mathcal{M}_{\mu}$ stabilization at gain $\mu$, and $\mathcal{D}_{\eta}$ stochastic disturbance of incoming amplitude $\eta$. In RF ordering, the controller update is available before exposure and attenuates the incoming disturbance:
\begin{align}
\mu_{\mathrm{RF}}^{+}(t)
&=\mathcal{C}\!\left(\rho(t),\mu(t)\right),\\
\psi_{\mathrm{nc}}^{\mathrm{RF}}(t)
&=\mathcal{D}_{\eta(1-\mu_{\mathrm{RF}}^{+}(t))}
\!\left[
\mathcal{M}_{\mu_{\mathrm{RF}}^{+}(t)}\bigl(\psi(t)\bigr)
\right],\\
\psi^{\mathrm{RF}}(t+\Delta t)
&=U_{H,\Delta t}\psi_{\mathrm{nc}}^{\mathrm{RF}}(t).
\end{align}
In DF ordering, disturbance reaches the state before a new regulatory response is available:
\begin{align}
\widetilde{\psi}^{\mathrm{DF}}(t)
&=\mathcal{D}_{\eta}\!\left(\psi(t)\right),\\
\mu_{\mathrm{DF}}^{+}(t)
&=\mathcal{C}\!\left(\widetilde{\rho}^{\mathrm{DF}}(t),\mu(t)\right),\\
\psi_{\mathrm{nc}}^{\mathrm{DF}}(t)
&=\mathcal{M}_{\mu_{\mathrm{DF}}^{+}(t)}
\!\left(\widetilde{\psi}^{\mathrm{DF}}(t)\right),\\
\psi^{\mathrm{DF}}(t+\Delta t)
&=U_{H,\Delta t}\psi_{\mathrm{nc}}^{\mathrm{DF}}(t).
\end{align}
The switching study does not redefine RF or DF. It changes only which already-defined ordering is applied at each integration step.

\section{Switching Protocols}
\label{sec:switching}

\begin{figure}[t]
\centering
\resizebox{0.90\linewidth}{!}{%
\begin{tikzpicture}[
    font=\small,
    rfblock/.style={draw=rfblue, fill=rfblue!16, minimum height=0.72cm, line width=0.65pt},
    dfblock/.style={draw=dforange, fill=dforange!16, minimum height=0.72cm, line width=0.65pt},
    axis/.style={-{Latex[length=2.5mm]}, line width=0.7pt},
    trans/.style={-{Latex[length=2.2mm]}, dashed, line width=0.7pt}
]
\node[anchor=east, font=\bfseries] at (-0.28,2.38) {Periodic:};
\foreach \x/\lab/\sty in {0/RF/rfblock,2/DF/dfblock,4/RF/rfblock,6/DF/dfblock,8/RF/rfblock} {
    \node[\sty, minimum width=2.0cm] at (\x+1,2.38) {\lab};
}
\draw[axis] (0,1.84) -- (10.72,1.84) node[right] {$t$};
\draw[decorate,decoration={brace,amplitude=5pt,mirror}, line width=0.7pt]
    (0,1.54) -- (2,1.54)
    node[midway,below=7pt] {dwell duration $L$};

\node[anchor=east, font=\bfseries] at (-0.28,-0.35) {Markov:};
\node[rfblock, minimum width=2.65cm] (mrfone) at (1.325,-0.35) {RF};
\node[dfblock, minimum width=1.65cm] (mdf) at (5.00,-0.35) {DF};
\node[rfblock, minimum width=2.90cm] (mrftwo) at (8.65,-0.35) {RF};
\draw[axis] (0,-0.92) -- (10.72,-0.92) node[right] {$t$};

\draw[trans] (mrfone.east) -- node[above=4pt, font=\scriptsize] {$p_{\mathrm{loss}}$} (mdf.west);
\draw[trans] (mdf.east) -- node[above=4pt, font=\scriptsize] {$p_{\mathrm{return}}$} (mrftwo.west);

\node[anchor=west, text=black!72, font=\scriptsize] at (0,-1.40)
    {Symmetric schedules: $p_{\mathrm{loss}}=p_{\mathrm{return}}$; episode duration varies.};
\end{tikzpicture}%
}
\caption{\textbf{Time-varying RF/DF protocols.} Periodic switching alternates RF and DF in fixed dwell blocks of length $L$. Stochastic switching uses a two-state Markov schedule with loss probability $p_{\mathrm{loss}}=P(\mathrm{RF}\rightarrow\mathrm{DF})$ and return probability $p_{\mathrm{return}}=P(\mathrm{DF}\rightarrow\mathrm{RF})$.}
\label{fig:schedule_schematic}
\end{figure}
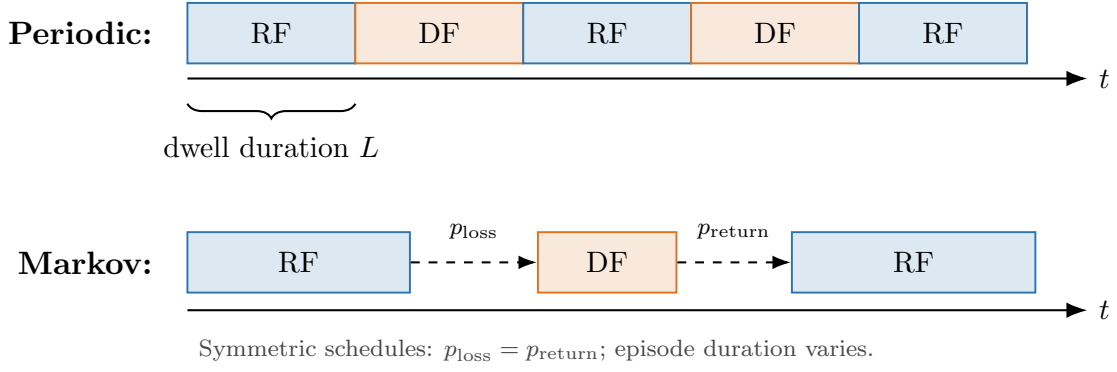

The central experimental variable is the time-dependent ordering
\begin{equation}
q(t)\in\{\mathrm{RF},\mathrm{DF}\}.
\end{equation}
At each integration step, the simulation driver evaluates $q(t)$ from a prescribed schedule, sets the agent's ordering, and then executes the corresponding RF or DF update. The periodic and Markov schedules are illustrated in Figure~\ref{fig:schedule_schematic}.

\subsection{Periodic switching}

Periodic switching alternates fixed RF and DF dwell blocks,
\begin{equation}
\mathrm{RF}^{L}\rightarrow\mathrm{DF}^{L}\rightarrow\mathrm{RF}^{L}\rightarrow\mathrm{DF}^{L}\rightarrow\cdots,
\label{eq:periodic}
\end{equation}
where $L$ is the number of integration steps spent in each mode. Equal-length RF and DF blocks give intended long-run RF occupancy $p_{\mathrm{RF}}=0.5$. Varying $L$ changes temporal persistence while holding overall mode balance fixed. The switching protocol acts on a continuously evolving regulatory state rather than on independent RF and DF samples: successive intervals inherit the controller state produced by the preceding intervals.

\subsection{Stochastic switching}

A two-state Markov schedule switches randomly between RF and DF. At each step, an RF trajectory loses RF access with probability $p_{\mathrm{loss}}$, while a DF trajectory regains RF access with probability $p_{\mathrm{return}}$. The next ordering depends only on the current ordering and these two probabilities, not on the earlier switching history. Thus,
\begin{equation}
P(\mathrm{RF}\rightarrow\mathrm{DF})=p_{\mathrm{loss}},
\qquad
P(\mathrm{DF}\rightarrow\mathrm{RF})=p_{\mathrm{return}}.
\label{eq:markov}
\end{equation}

The present analysis uses symmetric schedules,
\begin{equation}
p_{\mathrm{loss}}=p_{\mathrm{return}},
\end{equation}
so the expected RF occupancy is one-half, apart from finite-run fluctuations. The same probability also sets how long each episode tends to last. Measured in simulation steps, the expected episode duration is approximately
\begin{equation}
\langle L_{\mathrm{episode}}\rangle \approx \frac{1}{p_{\mathrm{loss}}}
= \frac{1}{p_{\mathrm{return}}},
\end{equation}
or, in simulation time,
\begin{equation}
\tau_{\mathrm{episode}}\approx
\frac{\Delta t}{p_{\mathrm{loss}}}
=
\frac{\Delta t}{p_{\mathrm{return}}}.
\end{equation}

A separate pseudorandom stream determines Markov switching. This keeps the switching decisions from changing the environmental-disturbance random sequence used in matched comparisons.

\section{Mixture Reference and Switching Penalty}
\label{sec:metrics}

\subsection{Occupancy-adjusted null model}

The null model is that only occupancy matters. A switching trajectory spending fraction $p_{\mathrm{RF}}$ of the analyzed interval in RF is compared with an occupancy-weighted mixture of matched fixed RF and fixed DF reference trajectories:
\begin{equation}
\overline{\mu}_{\mathrm{mix}}
=
 p_{\mathrm{RF}}\overline{\mu}_{\mathrm{RF}}
+
 \left(1-p_{\mathrm{RF}}\right)\overline{\mu}_{\mathrm{DF}}.
\label{eq:mixture}
\end{equation}
If temporal alternation has no effect beyond the time spent in each mode, then
\begin{equation}
\overline{\mu}_{\mathrm{switching}}\approx\overline{\mu}_{\mathrm{mix}}.
\end{equation}
Here $\overline{\mu}_{\mathrm{RF}}$ and $\overline{\mu}_{\mathrm{DF}}$ denote the post-burn-in time averages of the fixed-RF and fixed-DF reference trajectories \emph{matched to each switching replicate} (identical seed and environmental-disturbance stream), not pooled condition means. The switching penalty defined below is therefore a single paired quantity per replicate. The confidence intervals reported over the $N$ replicate-level penalties consequently incorporate the sampling variability of the reference trajectories, rather than treating $\overline{\mu}_{\mathrm{RF}}$ and $\overline{\mu}_{\mathrm{DF}}$ as fixed constants.

\subsection{Regulatory switching penalty}

The primary measure is
\begin{equation}
C_{\mathrm{switch}}^{(\mu)}
=
\overline{\mu}_{\mathrm{switching}}
-
\left[
 p_{\mathrm{RF}}\overline{\mu}_{\mathrm{RF}}
+
 \left(1-p_{\mathrm{RF}}\right)\overline{\mu}_{\mathrm{DF}}
\right].
\label{eq:penalty}
\end{equation}
A positive value indicates that switching imposes additional regulatory burden relative to the fixed-mode mixture. A negative value indicates regulatory relief: the switching trajectory requires less mean gain than predicted from fixed RF and fixed DF occupancy.

An analogous coherence-gap displacement is defined by replacing $\overline{\mu}$ with $\overline{\Delta C}$:
\begin{equation}
C_{\mathrm{switch}}^{(\Delta C)}
=
\overline{\Delta C}_{\mathrm{switching}}
-
\left[
 p_{\mathrm{RF}}\overline{\Delta C}_{\mathrm{RF}}
+
 \left(1-p_{\mathrm{RF}}\right)\overline{\Delta C}_{\mathrm{DF}}
\right].
\end{equation}
The present paper focuses on $C_{\mathrm{switch}}^{(\mu)}$ because it directly measures regulatory burden, the quantity in which the unexpected effect appears.

\subsection{Additional summaries and stationarity checks}

For each switching trajectory, the analysis also records post-burn-in averages of $\mu$ and $\Delta C$, the temporal susceptibility
\begin{equation}
\chi=\operatorname{Var}_{t\in\mathcal{T}_{\mathrm{post}}}\!\left[\Delta C(t)\right],
\end{equation}
and response amplitudes
\begin{equation}
A_{\mu}=\max_{t\in\mathcal{T}_{\mathrm{post}}}\mu(t)-\min_{t\in\mathcal{T}_{\mathrm{post}}}\mu(t),
\end{equation}
\begin{equation}
A_{\Delta C}=\max_{t\in\mathcal{T}_{\mathrm{post}}}\Delta C(t)-\min_{t\in\mathcal{T}_{\mathrm{post}}}\Delta C(t).
\end{equation}
For stochastic schedules, RF occupancy and mean DF episode duration are measured from the realized mode sequence.

A negative whole-run penalty could reflect a genuinely lower regulatory burden, or it could reflect a finite-time artifact in which $\mu(t)$ is still drifting upward and has not yet reached its late-time level. To check this, the post-burn-in interval is partitioned into four equal windows and the diagnostic
\begin{equation}
D_{\mu}=\overline{\mu}_{4}-\overline{\mu}_{3}
\end{equation}
is computed. A final-quarter linear slope $s_{\mu}$ is also estimated for $\mu(t)$. These diagnostics do not prove asymptotic convergence, but they test whether the observed relief is accompanied by unresolved upward regulatory accumulation over the analyzed horizon.

\section{Experimental Protocol}
\label{sec:protocol}

The high-statistics reference condition uses $\currentRuns$ matched replicates per condition, $\currentSteps$ integration steps per trajectory, and a burn-in of $\currentBurnIn$ steps. Matched fixed RF and fixed DF trajectories are used to construct the occupancy-adjusted reference for each switching replicate. The reference operating point is chosen in a regime where fixed DF requires greater mean gain than fixed RF, making it meaningful to ask how intermittent DF exposure changes regulatory burden.

\begin{table}[tb]
\centering
\caption{High-statistics RF/DF switching protocol used for the results reported in this manuscript.}
\label{tab:protocol}
\begin{tabular}{ll}
\toprule
Parameter & Value \\
\midrule
Current analyzed replicate count & $N=\currentRuns$ per switching condition \\
Steps per trajectory & \currentSteps{} \\
Burn-in discarded before analysis & \currentBurnIn{} steps \\
Time step $\Delta t$ & $0.01$ \\
State dimension $d$ & $16$ \\
Initial regulation gain $\mu_0$ & $0.05$ \\
Incoming disturbance amplitude $\eta$ & $0.17$ \\
Target entropy $S^*$ & $0.30$ \\
Hamiltonian energy scale & $0.15$ \\
Hamiltonian coupling $c$ & $0.08$ \\
Interaction locality $\ell$ & $2.0$ \\
Derivative feedback gain & $5\times10^{-4}$ \\
Target-error feedback gain & $2\times10^{-4}$ \\
Regulation bounds & $[0.001,1.0]$ \\
Periodic dwell durations $L$ & $1,10,50,100,500,1000,2000$ steps \\
Markov pairs $p_{\mathrm{loss}}:p_{\mathrm{return}}$ & $0.001{:}0.001$ to $0.05{:}0.05$ \\
Reference controls per seed & fixed RF and fixed DF \\
Primary statistic & $C_{\mathrm{switch}}^{(\mu)}$ \\
Uncertainty shown in primary plots & 95\% CI of condition mean \\
\bottomrule
\end{tabular}
\end{table}

The switching schedules include seven periodic dwell durations and six symmetric Markov loss/return probabilities. The same model parameters and analysis definitions are used across the fixed and switching conditions, so the only experimental change is the temporal schedule of RF and DF access.

\section{Results: Intermittent Control Produces Regulatory Relief}
\label{sec:results}

\subsection{All tested schedules show a negative switching penalty}

\begin{figure}[tb]
\centering
\includegraphics[width=0.98\linewidth]{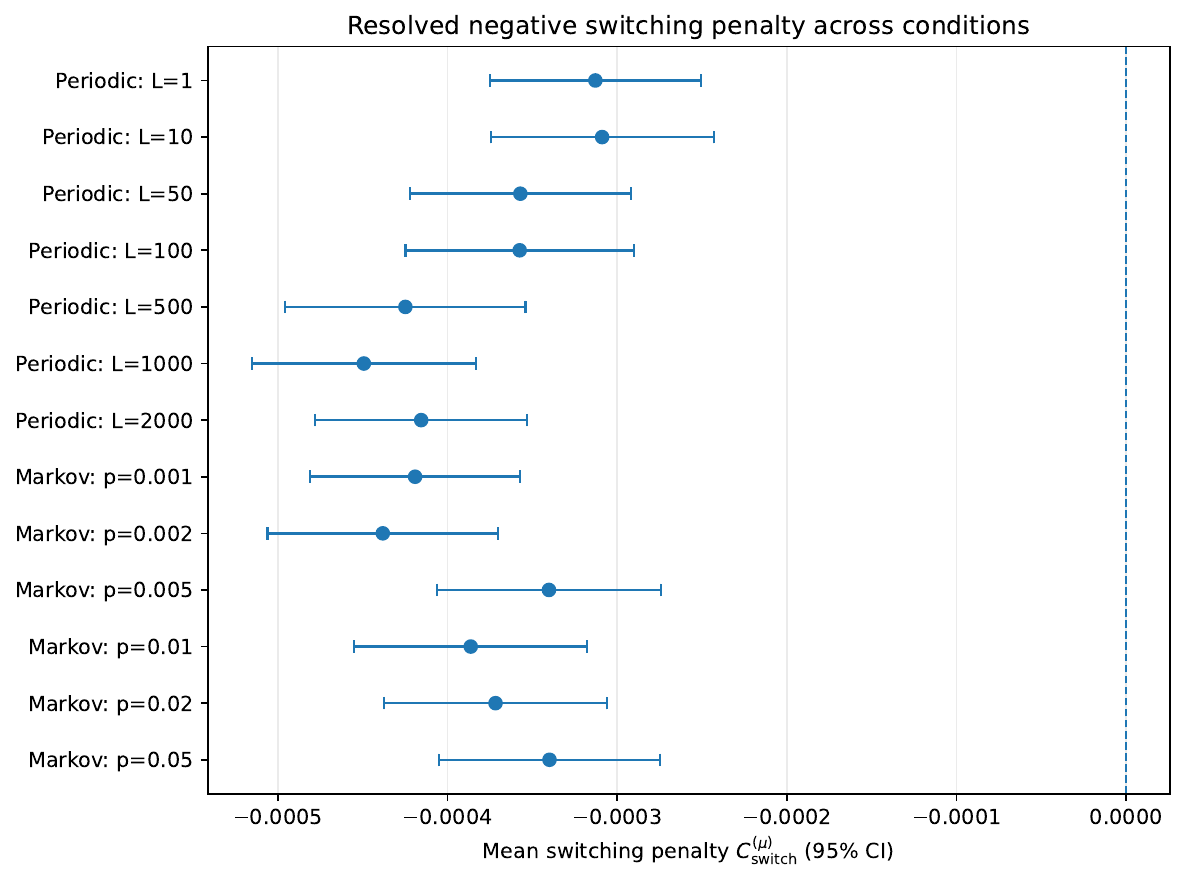}
\caption{\textbf{Robust negative nonlinear switching penalty across RF/DF schedules.}
Points report mean $C^{(\mu)}_{\mathrm{switch}}$, and horizontal bars show 95\% confidence intervals over $N=1000$ matched replicate-level penalties. The dashed vertical line marks the simple-mixture prediction $C^{(\mu)}_{\mathrm{switch}}=0$. For Markov schedules, $p=p_{\mathrm{loss}}=p_{\mathrm{return}}$. Every tested periodic and Markov condition lies below zero.}
\label{fig:forest}
\end{figure}

The main result is shown in Figure~\ref{fig:forest} and Table~\ref{tab:penalty_exact}. The initial expectation was a positive switching penalty: intermittent DF episodes would add burden beyond the fixed-mode mixture. Instead, every tested periodic and symmetric Markov schedule has negative $C_{\mathrm{switch}}^{(\mu)}$ with 95\% confidence intervals below zero.

\begin{table}[tb]
\centering
\small
\caption{\textbf{High-statistics nonlinear switching penalties.} Penalty means and 95\% confidence intervals are displayed in units of $10^{-4}$; negative values indicate less mean adaptive regulation than the occupancy-weighted fixed-RF/fixed-DF reference. $f_{-}$ is the fraction of replicate-level penalties below zero. All one-sided negative-penalty tests remain significant after false-discovery-rate adjustment.}
\label{tab:penalty_exact}
\begin{tabular}{llrrrr}
\toprule
Protocol & Condition & Mean DF duration & Mean & 95\% CI & $f_{-}$ \\
 & & (model time) & \multicolumn{2}{c}{$10^4 C_{\mathrm{switch}}^{(\mu)}$} & \\
\midrule
Periodic & $L=1$    & --    & $-3.128$ & $[-3.749,-2.507]$ & 0.631 \\
Periodic & $L=10$   & --    & $-3.088$ & $[-3.745,-2.431]$ & 0.611 \\
Periodic & $L=50$   & --    & $-3.570$ & $[-4.221,-2.920]$ & 0.656 \\
Periodic & $L=100$  & --    & $-3.574$ & $[-4.247,-2.901]$ & 0.654 \\
Periodic & $L=500$  & --    & $-4.247$ & $[-4.955,-3.540]$ & 0.647 \\
Periodic & $L=1000$ & --    & $-4.492$ & $[-5.152,-3.833]$ & 0.668 \\
Periodic & $L=2000$ & --    & $-4.154$ & $[-4.779,-3.530]$ & 0.676 \\
\midrule
Markov & $0.001/0.001$ & 9.999 & $-4.190$ & $[-4.811,-3.570]$ & 0.677 \\
Markov & $0.002/0.002$ & 5.002 & $-4.381$ & $[-5.060,-3.701]$ & 0.663 \\
Markov & $0.005/0.005$ & 2.001 & $-3.401$ & $[-4.061,-2.742]$ & 0.645 \\
Markov & $0.010/0.010$ & 1.000 & $-3.862$ & $[-4.548,-3.177]$ & 0.639 \\
Markov & $0.020/0.020$ & 0.500 & $-3.716$ & $[-4.373,-3.060]$ & 0.652 \\
Markov & $0.050/0.050$ & 0.200 & $-3.398$ & $[-4.050,-2.747]$ & 0.639 \\
\bottomrule
\end{tabular}
\end{table}

This result means that the switching agent does not behave like a simple average of fixed RF and fixed DF operation. Intermittent restoration of RF changes the regulatory burden associated with DF exposure. The finding should not be read as a claim that DF is beneficial in isolation. The fixed-ordering baselines still show that persistent DF requires greater regulation than persistent RF at the reference condition. The new result is that repeated switching between RF and DF can require less regulatory gain than predicted by mixing the persistent baselines.

In absolute terms the effect is small. At the reference operating point the matched fixed baselines average $\overline{\mu}_{\mathrm{RF}}=0.063$ and $\overline{\mu}_{\mathrm{DF}}=0.071$, so the occupancy-weighted mixture predicts $\overline{\mu}_{\mathrm{mix}}=0.067$, while the switching trajectories average $0.0669$. The mean penalties are therefore a few parts in $10^{4}$: roughly half a percent of the mean gain, or about five percent of the fixed RF-versus-DF separation itself. The fraction of replicates below zero, $f_{-}\approx0.63$--$0.68$ (Table~\ref{tab:penalty_exact}), shows that a substantial minority of individual runs still carry a positive penalty. The claim is therefore a robust downward shift in the \emph{mean} switching cost relative to the occupancy-weighted prediction, resolved by the high replicate count, rather than a large per-run effect.

Section~\ref{sec:discussion} considers a candidate mechanism---that an RF interval acts on burden inherited from the preceding DF episode rather than resetting the controller---together with the reasons this paper does not claim to have established it.

\subsection{Periodic switching is not explained by a single dwell duration}

\begin{figure}[tb]
\centering
\includegraphics[width=0.92\linewidth]{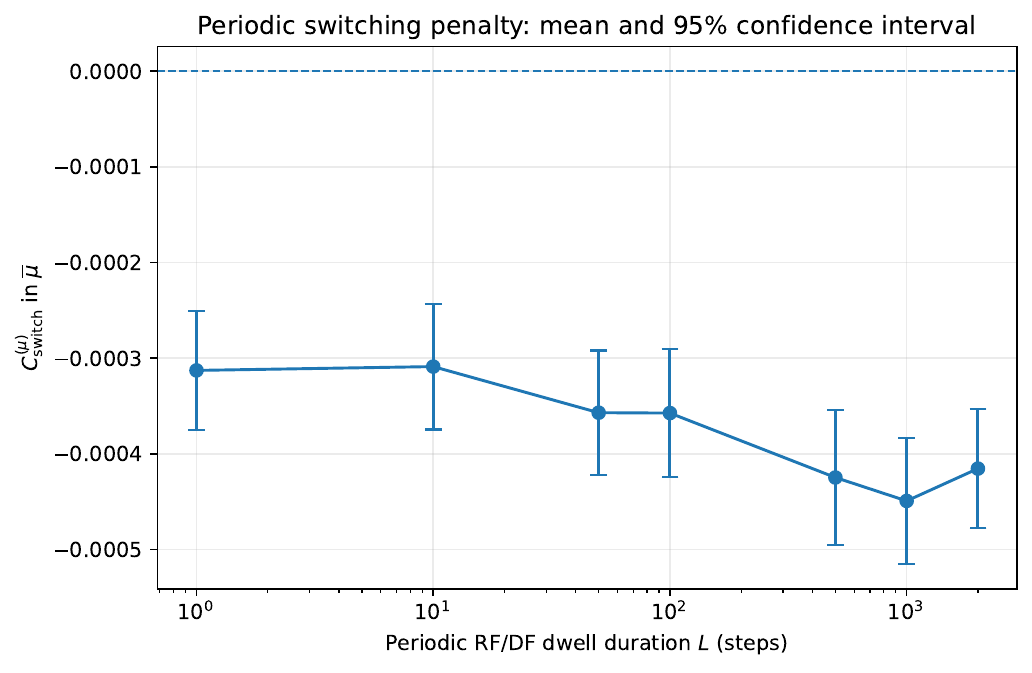}
\caption{\textbf{Periodic RF/DF switching penalty.}
Periodic schedules vary dwell duration $L$ while retaining equal RF and DF occupancy. All tested dwell durations show a negative mean penalty, with 95\% confidence intervals below zero.}
\label{fig:periodic_ci}
\end{figure}

Figure~\ref{fig:forest} establishes that every schedule is resolved below zero, but it orders conditions by label and therefore cannot show how the penalty varies with the switching timescale. Figures~\ref{fig:periodic_ci} and~\ref{fig:markov_ci} replot the same penalties against their controlling parameter---dwell duration $L$ and switch probability $p$---which is what the timescale arguments here and for the stochastic schedules below require.

The periodic protocol varies dwell duration while keeping RF and DF occupancy balanced. If the effect depended on one special timescale, one might expect only a narrow range of $L$ values to show relief. Instead, Figure~\ref{fig:periodic_ci} resolves a negative penalty at every tested dwell duration.

The magnitude is not identical across dwell lengths, indicating that temporal persistence matters. However, the sign of the effect is stable over the tested range. This supports the interpretation that the effect is a general switching phenomenon in the reference regime, not an accidental outcome of a single chosen dwell time.

\subsection{Aggregate periodic observables remain interpretable}

\begin{figure}[p]
\centering
\includegraphics[width=0.82\linewidth]{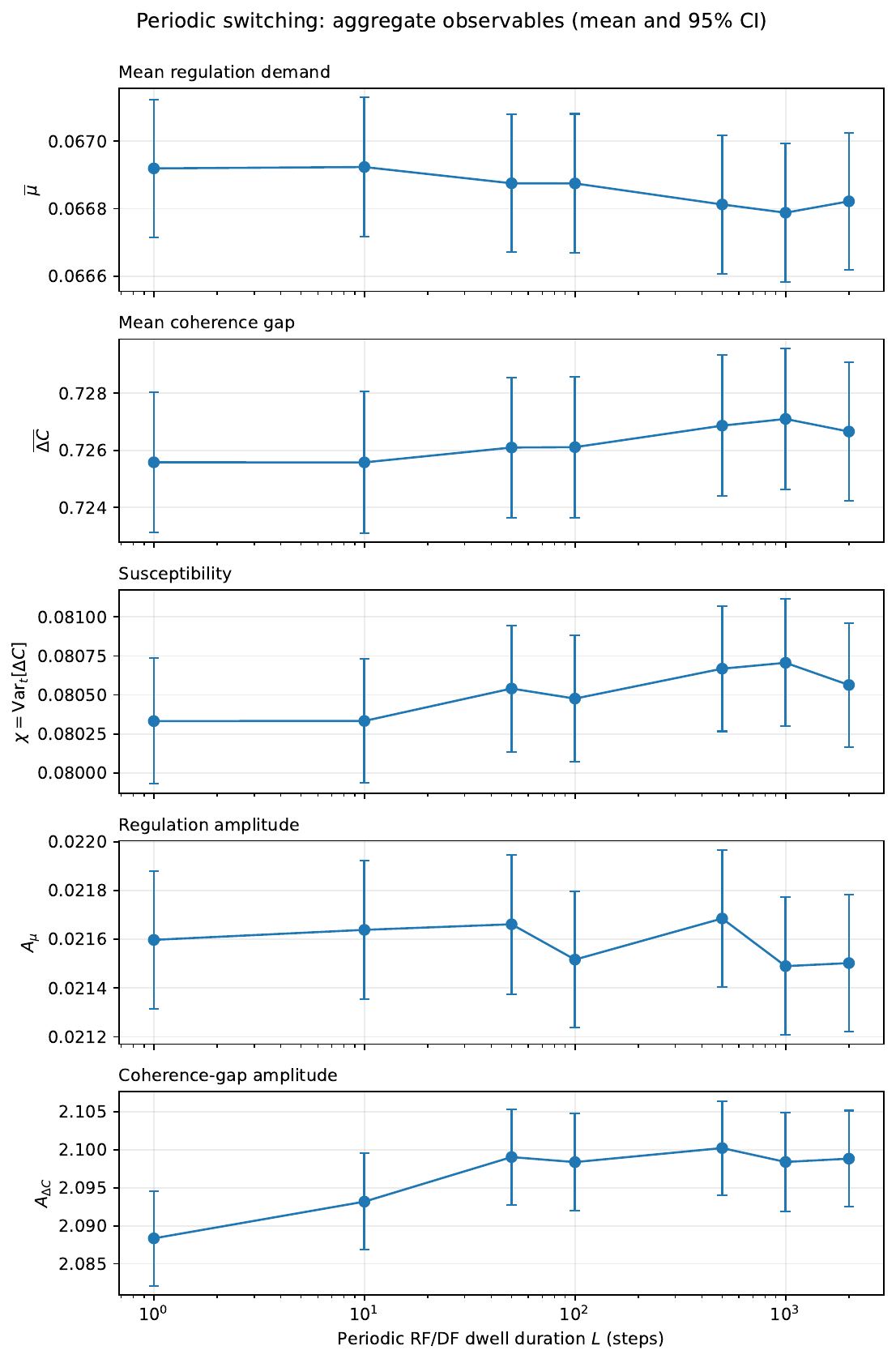}
\caption{\textbf{Aggregate observables under periodic switching.} Means and 95\% confidence intervals for regulation demand, coherence gap, susceptibility, regulation amplitude, and coherence-gap amplitude as a function of dwell duration $L$. These observables vary modestly across schedules while the nonlinear switching penalty remains robustly negative.}
\label{fig:periodic_metrics}
\end{figure}

Figure~\ref{fig:periodic_metrics} summarizes additional periodic observables. These quantities are not substitutes for the penalty calculation, but they help interpret what changes when dwell duration varies. The switching schedules alter mean regulatory burden, coherence-gap behavior, susceptibility, and response amplitudes in a coordinated way. The important point is that the primary regulatory relief appears without the other diagnostics behaving anomalously.

\subsection{Stochastic switching also produces regulatory relief}

\begin{figure}[tb]
\centering
\includegraphics[width=0.90\linewidth]{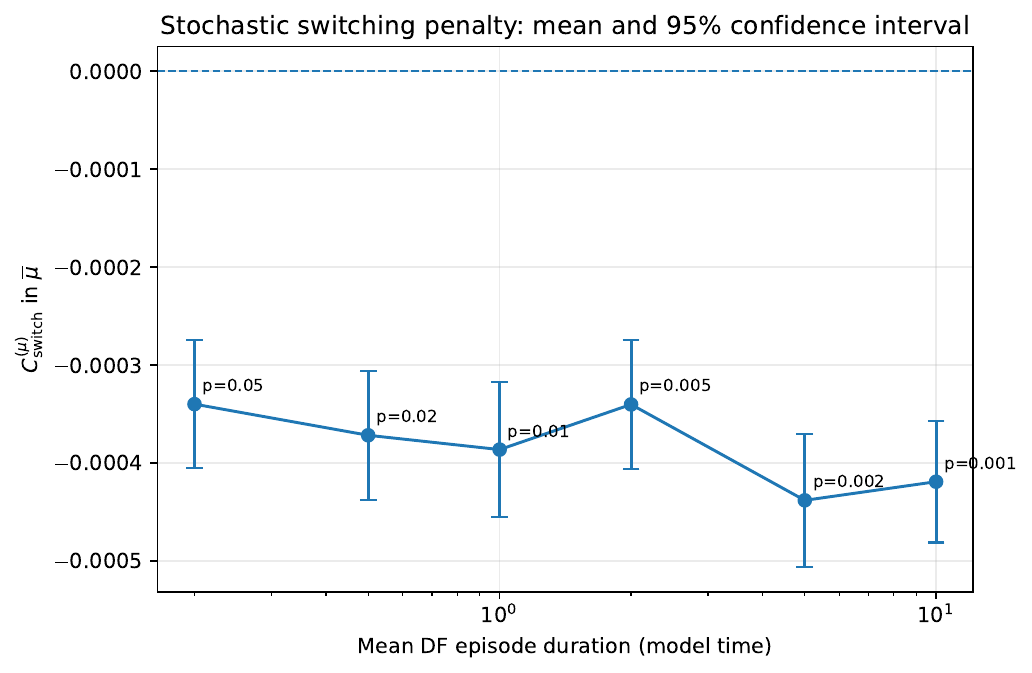}
\caption{\textbf{Stochastic RF/DF switching penalty.} Symmetric Markov schedules vary mean DF episode duration while retaining approximately equal expected RF and DF occupancy. Here $p=p_{\mathrm{loss}}=p_{\mathrm{return}}$, so smaller $p$ corresponds to longer expected episodes. All six tested schedules show a negative mean penalty with 95\% confidence intervals below zero.}
\label{fig:markov_ci}
\end{figure}

The negative penalty is not restricted to deterministic alternation. Figure~\ref{fig:markov_ci} shows the corresponding results for symmetric Markov schedules, again plotted against the controlling parameter rather than by condition label. These schedules vary mean episode duration stochastically while maintaining approximately balanced RF and DF occupancy. The sign remains negative across all tested probabilities.

The Markov result matters because it removes a possible artifact of perfectly regular alternation. Real agents are unlikely to lose and regain anticipatory access in exactly periodic blocks. The stochastic schedules show that the relief effect persists when switching times are irregular.

\subsection{Late-window diagnostics do not show unresolved upward accumulation}

\begin{figure}[tb]
\centering
\begin{subfigure}[t]{0.48\linewidth}
\centering
\includegraphics[width=\linewidth]{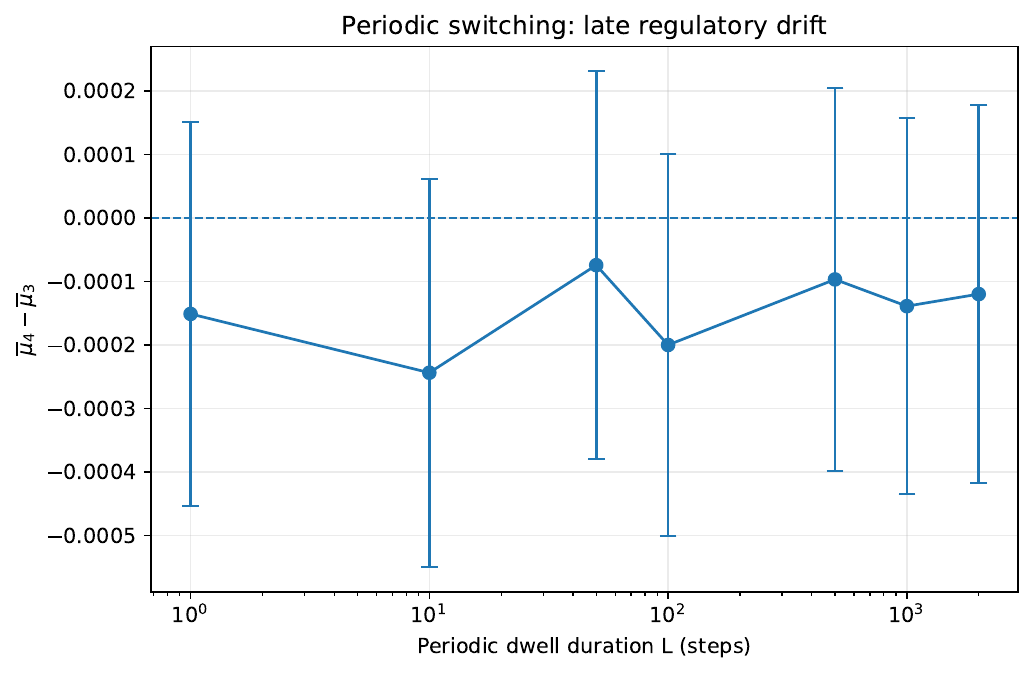}
\caption{Periodic: late-window drift $D_\mu$.}
\end{subfigure}\hfill
\begin{subfigure}[t]{0.48\linewidth}
\centering
\includegraphics[width=\linewidth]{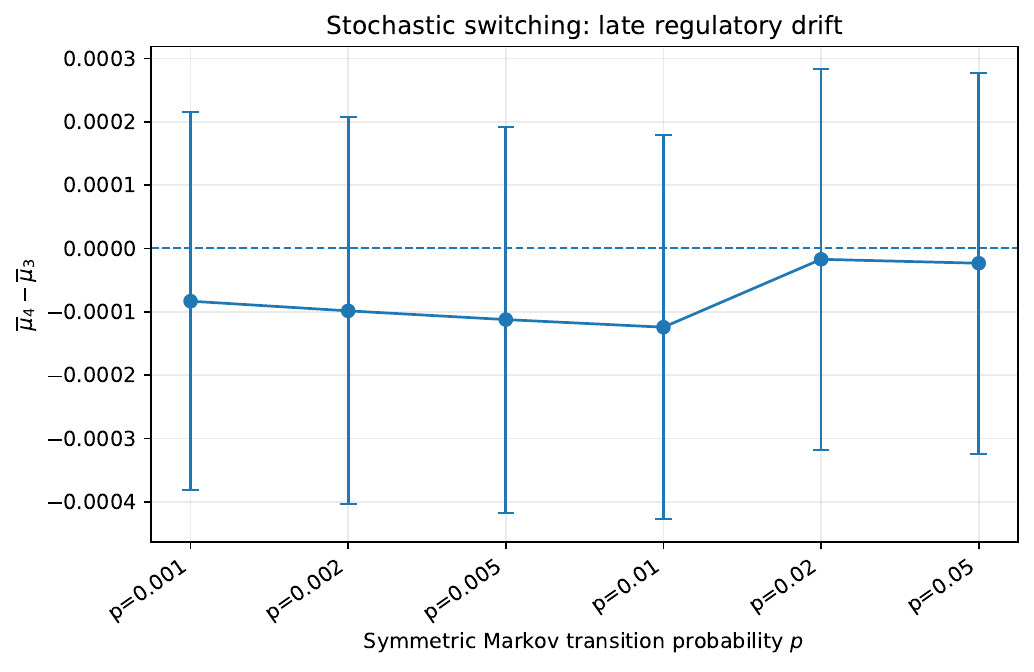}
\caption{Markov: late-window drift $D_\mu$.}
\end{subfigure}

\vspace{0.6em}

\begin{subfigure}[t]{0.48\linewidth}
\centering
\includegraphics[width=\linewidth]{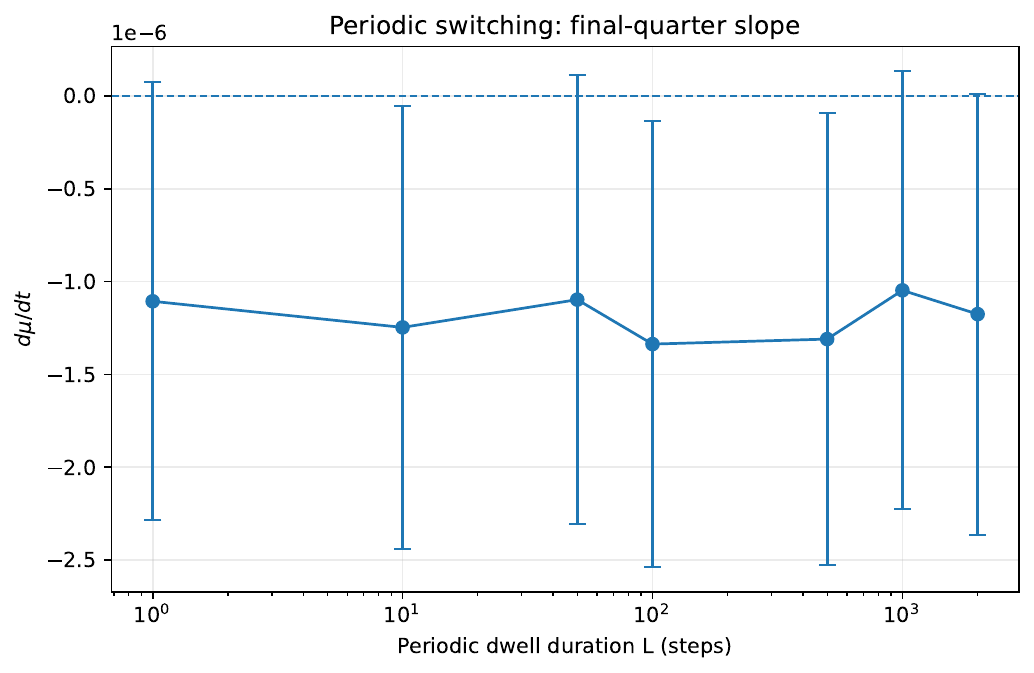}
\caption{Periodic: final-quarter slope $s_\mu$.}
\end{subfigure}\hfill
\begin{subfigure}[t]{0.48\linewidth}
\centering
\includegraphics[width=\linewidth]{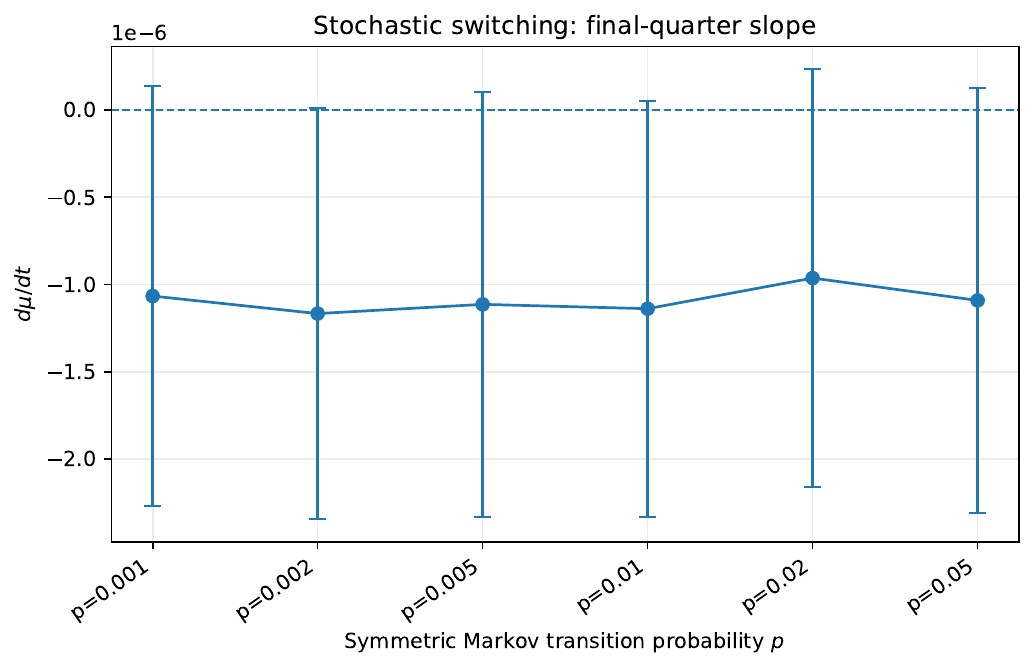}
\caption{Markov: final-quarter slope $s_\mu$.}
\end{subfigure}
\caption{\textbf{Late-time regulation diagnostics.}
Top panels show the change in mean gain between the fourth and third post-burn-in windows. Bottom panels show the linear slope of $\mu(t)$ over the final quarter of the trajectory. For Markov schedules, $p=p_{\mathrm{loss}}=p_{\mathrm{return}}$. Confidence intervals do not show resolved positive late-time accumulation under any tested switching schedule.}
\label{fig:stationarity}
\end{figure}

A concern with any long but finite simulation is that a low whole-run mean could occur because the trajectory has not yet accumulated the burden it would eventually reach. Figure~\ref{fig:stationarity} addresses this issue. The late-window diagnostics show no resolved evidence that $\mu(t)$ is still accumulating late in the run under any tested switching condition. Several periodic slopes are slightly negative.

These diagnostics do not establish an asymptotic theorem. They do, however, reduce the most direct finite-time objection to the main result. The observed negative penalty is not accompanied by evidence that $\mu(t)$ is still climbing upward over the analyzed horizon.

\section{Boundary Diagnostics: Where the Relief Effect Persists}
\label{sec:robustness}

The reference condition establishes the high-statistics result, but it does not by itself show whether the effect is local or broadly stable. The next analysis therefore asks how the switching penalty behaves over a bounded neighborhood of operating conditions. The goal is not to claim universal validity. It is to determine where the effect persists, where it attenuates, and whether attenuation becomes a genuine reversal.

\subsection{Bounded-domain protocol}

\begin{table}[t]
\centering
\small
\caption{\textbf{Bounded-domain robustness protocol.} The detailed $N=1000$ reference condition provides the principal high-statistics result; the added scans determine where the regulatory-relief effect persists or attenuates in nearby operating conditions.}
\label{tab:robustness_protocol}
\begin{tabular}{p{0.27\linewidth}p{0.61\linewidth}}
\toprule
Scan component & Values and representative switching protocols \\
\midrule
Regulation/noise map &
$\mu_0 \in \{0.03,0.04,0.05,0.06,0.07\}$,
$\eta \in \{0.13,0.15,0.17,0.19,0.21\}$,
with $S^*=0.30$ \\
Target-uncertainty sensitivity &
$S^* \in \{0.20,0.25,0.30,0.35,0.40\}$,
with $(\mu_0,\eta)=(0.05,0.17)$ \\
Periodic protocols &
$L=500$ and $L=1000$ steps \\
Stochastic protocols &
$p_{\mathrm{loss}}=p_{\mathrm{return}}=0.002$ and $0.001$ \\
Statistics per point &
$N=200$ matched replicates, $200{,}000$ steps, $40{,}000$-sample burn-in \\
\bottomrule
\end{tabular}
\end{table}

The bounded scan varies initial gain, disturbance level, and target uncertainty. For each condition, fixed RF and fixed DF baselines are compared with switching trajectories. This allows the switching penalty to be interpreted relative to the local fixed-ordering separation rather than relative only to the original reference point.

\subsection{Fixed-ordering separation and switching relief}

A useful boundary diagnostic is the local fixed-ordering separation
\begin{equation}
B_{\mathrm{RF/DF}}^{(\mu)}=
\overline{\mu}_{\mathrm{DF}}-
\overline{\mu}_{\mathrm{RF}}.
\end{equation}
Positive $B_{\mathrm{RF/DF}}^{(\mu)}$ indicates a local RF advantage in regulatory burden: persistent DF requires more mean gain than persistent RF. The switching penalty can then be evaluated in relation to this local baseline separation.

\begin{figure}[p]
\centering
\includegraphics[width=0.99\linewidth]{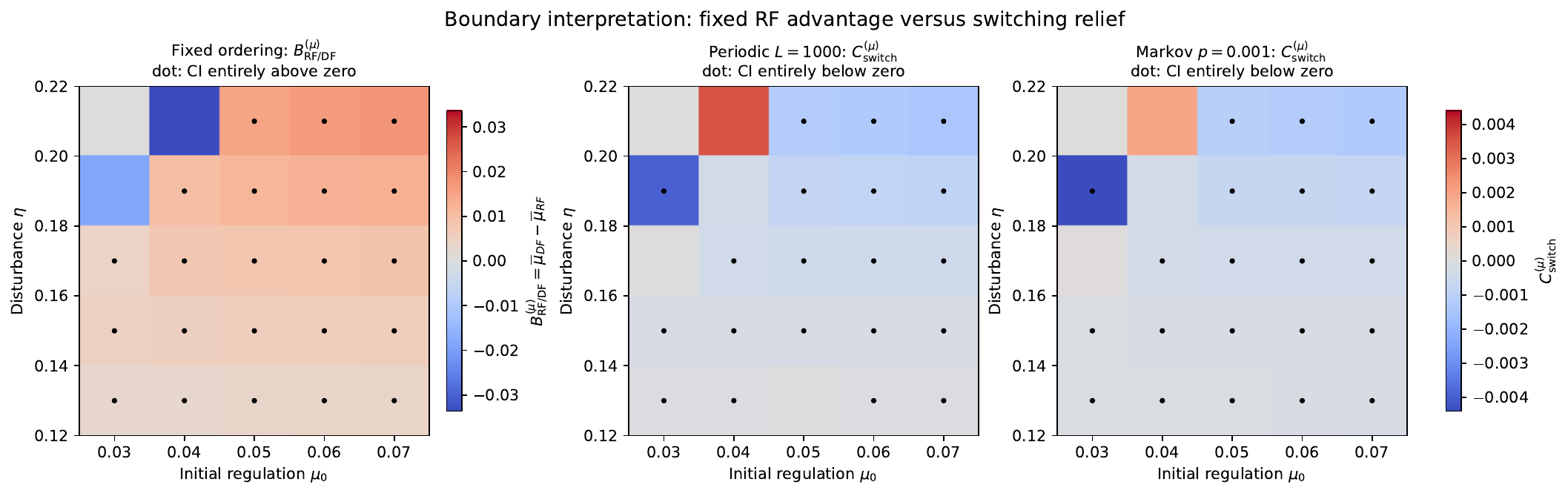}
\caption{\textbf{Fixed-ordering advantage and switching relief over the bounded $(\mu_0,\eta)$ scan.} Left: the local fixed-ordering separation $B_{\mathrm{RF/DF}}^{(\mu)}=\overline{\mu}_{\mathrm{DF}}-\overline{\mu}_{\mathrm{RF}}$; black dots indicate 95\% confidence intervals entirely above zero, so fixed RF has resolved lower regulatory burden than fixed DF. Center and right: $C_{\mathrm{switch}}^{(\mu)}$ for representative periodic and stochastic switching schedules; black dots indicate 95\% confidence intervals entirely below zero. Resolved switching relief co-occurs with fixed RF advantage through most of the tested domain, whereas attenuation is concentrated near the boundary at low initial regulation and high disturbance.}
\label{fig:phase_baseline_diagnostic}
\end{figure}

Figure~\ref{fig:phase_baseline_diagnostic} shows that the switching relief extends beyond the single high-statistics reference condition. At the same time, the magnitude of the effect is not uniform. It attenuates near parts of the bounded scan where the fixed RF-versus-DF separation itself becomes small or unresolved. This is a crucial distinction: attenuation of relief is not the same as evidence that switching has become costly in a regime where RF remains clearly advantageous.

\subsection{No resolved reversal inside the RF-advantage regime}

\begin{figure}[p]
\centering
\includegraphics[width=0.94\linewidth]{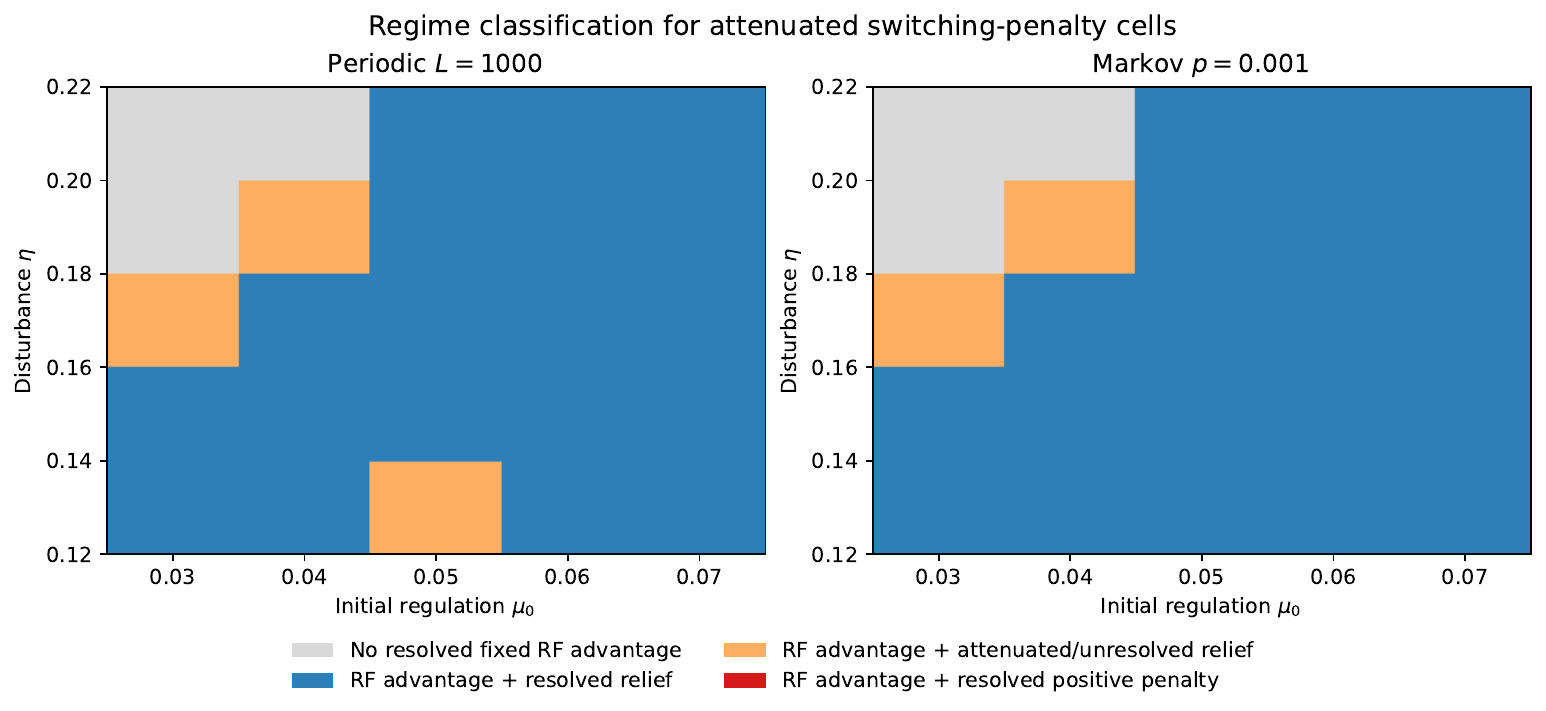}
\caption{\textbf{Joint classification of the bounded phase-map cells.} Blue cells show resolved fixed RF advantage together with resolved negative switching penalty. Amber cells retain resolved RF advantage but have attenuated or statistically unresolved switching relief. Gray cells have no resolved fixed RF advantage. No red cells occur in the scanned domain: the scan provides no evidence for a resolved positive switching penalty inside a region retaining resolved RF advantage.}
\label{fig:phase_regime_classification}
\end{figure}

Figure~\ref{fig:phase_regime_classification} directly classifies scanned cells according to both fixed-ordering separation and switching penalty. The key boundary result is a null one: no tested cell combines a resolved fixed RF advantage with a resolved positive switching penalty.

This matters because a positive switching penalty inside the RF-advantage regime would support the original burden hypothesis: intermittent loss of RF would add regulatory burden beyond the fixed-mode mixture. The scan does not show that pattern. Instead, the result is relief where the effect is resolved and attenuation where the local RF/DF separation weakens.

\subsection{Target uncertainty attenuates both fixed-ordering advantage and switching relief}

\begin{figure}[tb]
\centering
\includegraphics[width=0.74\linewidth]{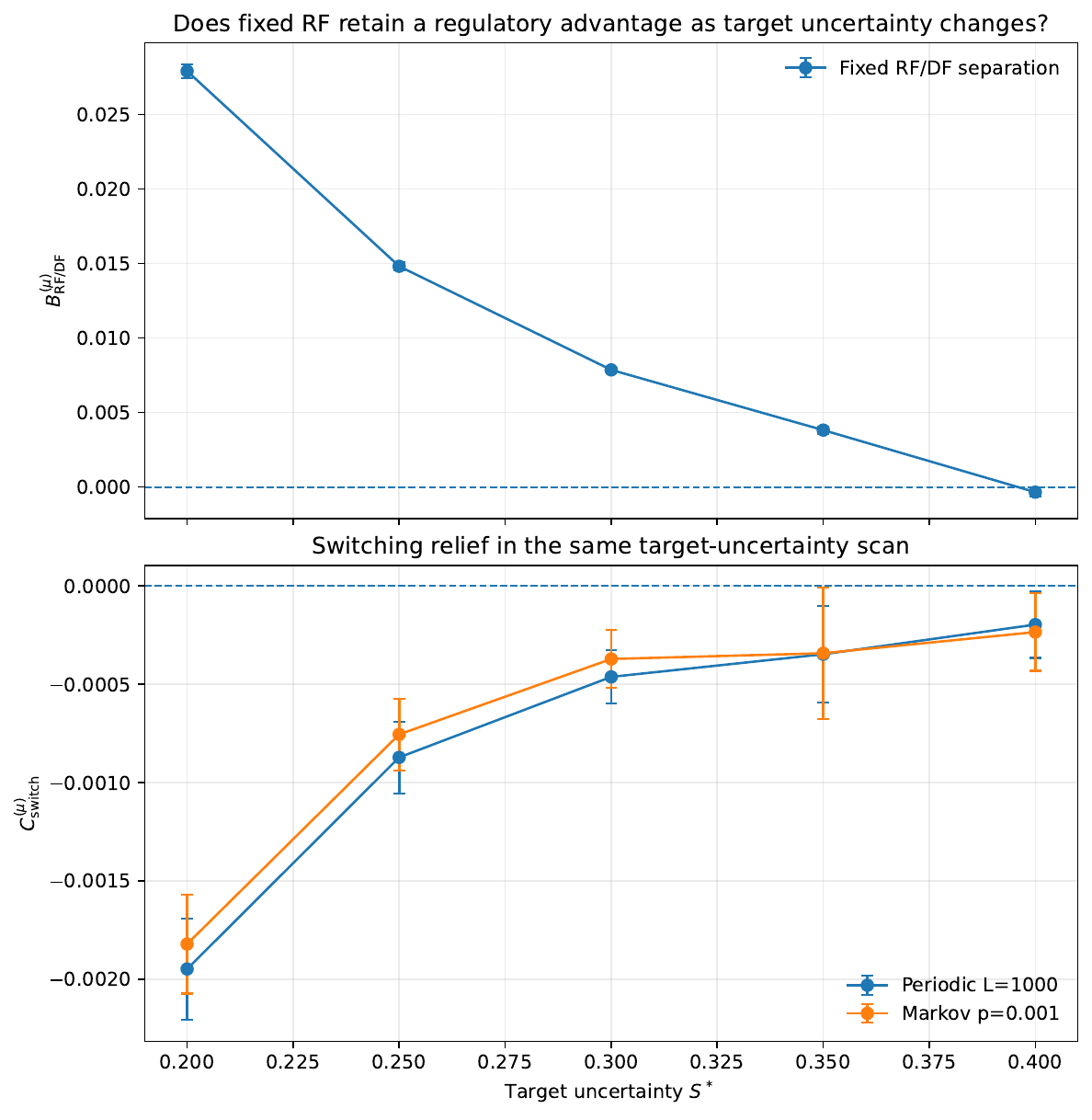}
\caption{\textbf{Fixed-ordering separation and switching relief across target uncertainty.} Top: $B_{\mathrm{RF/DF}}^{(\mu)}$ decreases as $S^*$ increases and reaches approximately zero at the highest tested target uncertainty. Bottom: the corresponding switching penalties for representative periodic and stochastic schedules remain negative in mean but attenuate toward zero. These paired results show that high-$S^*$ attenuation accompanies erosion of the fixed RF-versus-DF advantage.}
\label{fig:target_baseline_diagnostic}
\end{figure}

The target-uncertainty scan provides an additional boundary check. As $S^*$ increases, the difference between fixed RF and fixed DF operation decreases. Figure~\ref{fig:target_baseline_diagnostic} shows that switching relief attenuates in parallel with this collapse of fixed-ordering separation.

This pattern supports the interpretation that the switching effect depends on a regime in which causal ordering creates a meaningful regulatory-burden separation. When the fixed RF/DF distinction no longer produces a resolved burden difference, the nonlinear switching relief also becomes unresolved.

\section{Discussion}
\label{sec:discussion}

\subsection{Interpretation and significance}

The central result is not simply that RF is better than DF. That was already suggested by fixed-ordering comparisons. The new result is that intermittent RF/DF access is not explained by the fraction of time spent in each mode. If occupancy alone controlled the burden, the mean gain of a switching trajectory would match the weighted average of the fixed-RF and fixed-DF baselines. Instead, the switching trajectories require less mean gain than that prediction. The switching history therefore changes the subsequent regulatory burden.

This matters for artificial-agent evaluation because many adaptive systems are exposed to intermittent degradation rather than permanent regime changes. An agent may briefly lose clean perception, operate under delayed input, enter a noisy environment, or shift between high- and low-resource modes. Fixed-mode tests remain necessary, but they cannot answer whether temporary loss followed by recovery has the same burden as diluted permanent loss. The present result shows that, at least in this model, the answer is no.

\subsection{A candidate dynamical interpretation}

The result suggests a trajectory-level mechanism, of the kind familiar from nonlinear and self-organizing dynamical systems~\cite{strogatz2015nonlinear,kelso1995dynamic}. In persistent DF operation, disturbance repeatedly enters before a new regulatory response can act, producing a higher mean gain. In switching operation, DF episodes are embedded within repeated returns to RF. Those RF intervals may prevent DF-associated burden from accumulating as though the system were permanently reactive. The agent is therefore not merely spending some fraction of time in DF; it is repeatedly re-entering a causal regime that changes how subsequent disturbance is encountered.

The effect also has an equivalent static description. A negative $C_{\mathrm{switch}}^{(\mu)}$ at balanced occupancy is the same statement as the mean gain being a convex function of RF occupancy: the switching trajectory lies below the chord joining the fixed-RF and fixed-DF endpoints. Because balanced occupancy can only be realized \emph{through} switching---there is no static intermediate-occupancy agent to compare against---this convexity and the temporal switching effect are two descriptions of the same mean-level measurement.

This interpretation therefore remains a hypothesis about mechanism. The present paper establishes the effect, checks stationarity, and maps a bounded operating region; distinguishing a genuinely dynamic, history-inheriting mechanism from a static occupancy--gain nonlinearity would require the event-aligned analysis of RF-to-DF and DF-to-RF transitions reserved for subsequent work.

\subsection{Relevance beyond the model}

IRAM-$\Omega$-Q is useful in this context because it isolates a feature that many adaptive agents share: the agent has an internal state, a stabilizing controller, a disturbance process, and memory of prior evolution. The model is not intended to reproduce every detail of deployed AI systems. Its value is that the timing of regulation and disturbance can be changed while the rest of the dynamics are held fixed. This makes it possible to separate an occupancy effect from a history-dependent switching effect.

The broader interpretation is therefore not that all artificial agents will show the same numerical pattern. Rather, the result shows that fixed-mode evaluation can miss a class of timing-dependent behavior. In any adaptive system whose internal state carries the consequences of earlier disruption and recovery, intermittent degradation may produce a regulatory burden that differs from the weighted average of permanently available and permanently unavailable control. IRAM-$\Omega$-Q provides a transparent test case in which this effect can be measured directly, making it a useful guide for designing comparable tests in other artificial-agent architectures.

\subsection{Implications for adaptive-agent design}

The results suggest several design-relevant lessons.

First, intermittent degradation should be evaluated dynamically. The cost of losing anticipatory access may depend on how long the loss lasts, how recovery occurs, and how much state history is retained.

Second, recovery access is itself a design variable. Surviving degraded operation may not be enough: a robust design should also restore anticipatory control quickly enough to keep reactive burden from accumulating.

Third, internal regulatory burden can complement external task-performance measures. Two agents may perform similarly while requiring different levels of internal correction. In practical architectures, an analogue of $\mu$ could be corrective computation, uncertainty gating, repeated replanning, safety-monitor activation, or control effort.

Fourth, fixed-mode benchmarks can miss nonlinear effects of control timing. If a system is intended to operate under fluctuating sensing, communication, or compute conditions, switching protocols should be part of the evaluation suite.

\section{Scope and Limitations}
\label{sec:limitations}

The result should be interpreted within four boundaries.

\begin{enumerate}
    \item \textbf{Model-specific burden.} The gain $\mu(t)$ is an operational measure inside IRAM-$\Omega$-Q. It motivates analogous burden measures in other agents, but it is not a universal measure of energy, computation, or deployed-system cost.

    \item \textbf{Bounded operating region.} The scans characterize a defined neighborhood of the reference condition. They do not establish that intermittent anticipatory control is beneficial under all parameters, controllers, disturbance models, or architectures.

    \item \textbf{Finite-time stationarity diagnostics.} The late-window diagnostics reduce concern about unresolved upward accumulation over the analyzed horizon, but they are not a proof of asymptotic convergence.

    \item \textbf{Mechanism not fully resolved.} The paper identifies a robust nonlinear switching effect and maps its boundary. Transition-level response analysis is left for subsequent mechanism-focused work.
\end{enumerate}

The boundary diagnostic is also limited to the scanned domain. No tested cell shows a resolved positive switching penalty inside a resolved RF-advantage regime, but this is not a universal impossibility theorem.

\section{Conclusion}

This paper tested a simple fixed-mode prediction: an adaptive agent alternating between anticipatory and reactive control should require the occupancy-weighted average of the two fixed-mode burdens. In IRAM-$\Omega$-Q, the switching trajectories violate that prediction.

At the high-statistics reference condition, all tested periodic and symmetric Markov switching schedules produce a negative nonlinear switching penalty in adaptive gain. The switching agent requires less mean regulation than predicted by a matched occupancy-weighted mixture of fixed RF and fixed DF trajectories. Late-window diagnostics do not show unresolved upward accumulation of regulatory gain, and bounded scans show that the relief effect persists over a neighborhood of the reference regime.

The boundary diagnostics clarify the limit of the effect. Relief attenuates where the fixed RF-versus-DF burden separation becomes unresolved or collapses toward zero. Within the tested region, no cell combines a resolved fixed RF advantage with a resolved positive switching penalty. The evidence therefore supports a regulatory-relief regime rather than a hidden intermittent-burden regime inside the resolved RF-advantage domain.

The broader implication is methodological. Adaptive agents should not be evaluated only by fixed-regime performance when their operating conditions are intermittent. The sequence of disturbance, response, and recovery can shape the internal burden required to remain organized. IRAM-$\Omega$-Q provides one reproducible case in which intermittent restoration of anticipatory control produces a nonlinear benefit relative to fixed-mode averaging.

\section*{Data and Code Availability}

The simulations and analyses were generated using the IRAM-$\Omega$-Q Java framework and associated analysis scripts. Configuration files, analysis scripts, and the numerical summary tables underlying every reported figure are available from the author on request, and will accompany the public release of the framework code.

\end{document}